\PassOptionsToPackage{table}{xcolor}

\documentclass[sigconf]{acmart}
\usepackage[table]{xcolor}
\usepackage{multicol}
\usepackage{multirow}
\usepackage{tcolorbox}
\tcbuselibrary{skins, breakable, theorems}

\copyrightyear{2024}
\acmYear{2024}

\acmConference[]{Preprint}{Under Review,
  2024}{}
\renewcommand\footnotetextcopyrightpermission[1]{}

\author{Kai Xiong}
\email{kxiong@ir.hit.edu.cn}
\affiliation{
  \institution{Harbin Institute of Technology}
  \city{Harbin}
  \country{China}
}

\author{Xiao Ding\authornotemark[1]}
\email{xding@ir.hit.edu.cn}
\affiliation{
  \institution{Harbin Institute of Technology}
  \city{Harbin}
  \country{China}
}

\authornote{Corresponding Authors}

\author{Li Du}
\email{duli@baai.ac.cn}
\affiliation{
  \institution{Beijing Academy of Artificial Intelligence}
  \city{Beijing}
  \country{China}
}

\author{Jiahao Ying}
\email{jhying.2022@phdcs.smu.edu.sg}
\affiliation{
  \institution{Singapore Management University}
  \country{Singapore}
}

\author{Ting Liu}
\email{tliu@ir.hit.edu.cn}
\affiliation{
  \institution{Harbin Institute of Technology}
  \city{Harbin}
  \country{China}
}

\author{Bing Qin}
\email{qbin@ir.hit.edu.cn}
\affiliation{
  \institution{Harbin Institute of Technology}
  \city{Harbin}
  \country{China}
}
\author{Yixin Cao}
\email{yxcao@fudan.edu.cn}
\authornotemark[1]
\affiliation{
  \institution{Fudan University}
  \city{Shanghai}
  \country{China}
}

\begin{document}

\title{Diagnosing and Remedying Knowledge Deficiencies in LLMs via Label-free Curricular Meaningful Learning}


\renewcommand{\shortauthors}{Xiong et al., 2024}

\begin{abstract}

  Large Language Models~(LLMs) are versatile and demonstrate impressive generalization ability by mining and learning information from extensive unlabeled text. However, they still exhibit reasoning mistakes, often stemming from knowledge deficiencies, which can affect their trustworthiness and reliability. Although users can provide diverse and comprehensive queries, obtaining sufficient and effective feedback is demanding. Furthermore, evaluating LLMs comprehensively with limited labeled samples is difficult. This makes it a challenge to diagnose and remedy the deficiencies of LLMs through rich label-free user queries. To tackle this challenge, we propose a label-free curricular meaningful learning framework~(LaMer). LaMer first employs relative entropy to automatically diagnose and quantify the knowledge deficiencies of LLMs in a label-free setting. Next, to remedy the diagnosed knowledge deficiencies, we apply curricular meaningful learning: first, we adopt meaningful learning to adaptively synthesize augmentation data according to the severity of the deficiencies, and then design a curricular deficiency remedy strategy to remedy the knowledge deficiencies of LLMs progressively. Experiments show that LaMer efficiently and effectively diagnoses and remedies knowledge deficiencies in LLMs, improving various LLMs across seven out-of-distribution~(OOD) reasoning and language understanding benchmarks, achieving comparable results to baselines with just 40\% training data. LaMer even surpasses methods that rely on labeled datasets for deficiency diagnosis. In application, our label-free method can offer an effective knowledge deficiency diagnostic tool for efficient LLM development.
\end{abstract}

\maketitle

\section{Introduction}

Large language models (LLMs) have made significant advancements in various fields recently~\cite{kim2024large,li2024common,kim2024financial}. By implicitly mining and learning information from vast amounts of unlabeled text via language modeling, LLMs have demonstrated remarkable generalization abilities. This enables them to answer a wide array of user queries across many applications such as healthcare~\cite{lungren2024more,tripathi2024efficient} and recommender systems~\cite{lei2023recexplainer,wu2024coral}. However, despite their potential, LLMs still have limitations. Due to their statistical nature, LLMs occasionally make reasoning mistakes~\cite{jung2022maieutic,wang2023large}, which can undermine user trust and the reliability of their applications. A significant challenge is that the knowledge mining process is implicit, making it difficult to discern what LLMs are particularly good or bad at. This lack of transparency hinders targeted improvements and quality assurance of LLMs. Additionally, relying on users for sufficient and effective feedback is often difficult and impractical, as it requires extra effort and users typically seek answers to questions they do not fully understand. This situation poses a significant obstacle in continually improving LLMs based on massive label-free user queries.

\begin{figure}[t]
    \centering
    \includegraphics[width=0.9\linewidth]{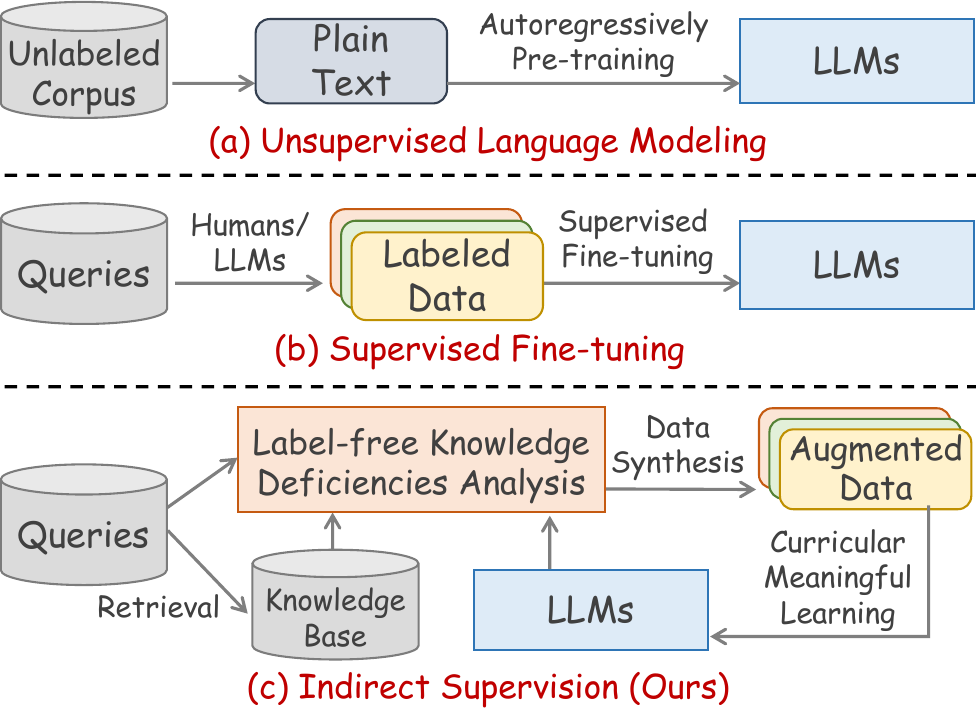}
    \caption{Differences among (a)~unsupervised language modeling, (b)~supervised fine-tuning, and (c) our proposed indirect supervision method.}
    \label{fig:intro}
    \Description{Simple examples to show differences among different methods.}
\end{figure}

To enhance the performance of LLMs, current researches predominantly follow two methodologies: unsupervised language modeling~\cite{fujii2024continual,guo2024efficient} and supervised fine-tuning~(SFT)~\cite{mitra2023orca,xu2024wizardlm}, which is respectively shown in Figure~\ref{fig:intro} (a) and (b). Unsupervised language modeling leverages vast quantities of unlabeled text data, enabling LLMs to learn knowledge implicitly from the enormous volume of available information. In contrast, SFT involves training LLMs on labeled datasets tailored to specific tasks. Despite their advantages, both approaches have limitations. First, they can be inefficient as it necessitates the inclusion of extensive data indiscriminately, and may not address enough reasoning mistakes of LLMs. Second, they still lack a comprehensive understanding of LLMs, leading to the inability to make targeted improvements. As a result, this may further lead to ineffectiveness in addressing specific and long-tail user demands. Moreover, labeling user queries can reveal some reasoning mistakes of LLMs, while it is costly and challenging to use limited labeled testing samples to comprehensively evaluate such powerful LLMs with good generalization capability~\cite{liu2023g,zheng2024judging}. These limitations underscore the necessity for more efficient and cost-effective methods to diagnose and improve existing LLMs.


To tackle the above challenges, and considering that reasoning mistakes in LLMs often arise from knowledge deficiencies such as lack of knowledge and ineffective application of existing knowledge~\cite{gao2023retrieval,xiong2024meaningful}, we aim to diagnose the knowledge deficiencies in LLMs and remedy them without any costly annotations. By diagnosing these knowledge deficiencies, we can tailor solutions for targeted and efficient improvements. Furthermore, our method enables LLMs to evolve continuously with increasing user engagement, even in the absence of user feedback. This approach ensures that LLMs remain current and adaptable, capable of handling specific and long-tail user demands, avoiding costly data labeling processes, and promoting a more efficient development cycle.

In this paper, as shown in Figure~\ref{fig:intro} (c), we design an indirect supervision method called label-free curricular meaningful learning~(LaMer), which first diagnoses the knowledge deficiencies of a specific LLM in a label-free setting, and then devise curricular meaningful learning to efficiently and effectively remedy the deficiencies of corresponding LLM. Specifically, LaMer first retrieves relevant knowledge for user queries. Subsequently, inspired by the information theory~\cite{shannon1948mathematical}, where relative entropy~\cite{kullback1951information} can estimate the extra information needed to change from a distribution to another distribution. We leverage relative entropy with the retrieved knowledge to diagnose the knowledge deficiencies of LLMs without relying on labels. Finally, we adopt curricular meaningful learning, which initially uses meaningful learning~\cite{xiong2024meaningful} to adaptively synthesize augmented data across various scenarios based on the severity of the deficiencies, and then employ curricular deficiency remedy to progressively address these deficiencies from minor to severe.

We conduct extensive experiments on 4 open-source LLMs and evaluate LaMer and baselines across 7 out-of-distribution~(OOD) reasoning and language understanding benchmarks. The results show that LaMer proficiently diagnoses the knowledge deficiencies in various LLMs, leading to more efficient and effective improvements compared to baselines. It achieves comparable results to baselines with just 40\% training data. Further analyses reveal that LaMer not only surprisingly surpasses methods relying on labeled data to detect deficiencies but also highlights its efficiency and effectiveness in diagnosing and remedying knowledge deficiencies in LLMs, offering a robust solution for improving their application.

We summarize our contributions as follows:
\begin{itemize}
    \item We incorporate relative entropy to automatically and effectively diagnose knowledge deficiencies in existing LLMs without relying on labeled data, thus breaking the limitation of relying solely on existing labeled datasets.
    \item We develop curricular meaningful learning to efficiently and proficiently remedy the knowledge deficiencies in LLMs.
    \item Our proposed LaMer excels in diagnosing and remedying the knowledge deficiencies, towards maximizing the potential of existing open-source LLMs.
\end{itemize}

\section{Related Work}
\subsection{Continual Training on LLMs}
LLMs~\cite{jiang2023mistral,llama3modelcard} have demonstrated strong generalization capabilities, but they may not always perform satisfactorily on general or specific capabilities of interest to individuals. Recent works involved additional training to customize or enhance LLMs.

For enhancing the general capabilities of LLMs, \citet{xie2023efficient} and \citet{li2024examining} conducted continual pre-training on existing LLMs for alignment. \citet{cui2023efficient} and \citet{huozi} conducted adaptions with massive Chinese data to enhance the Chinese capabilities of LLaMA~\cite{touvron2023llama}. \citet{alpaca} utilized 52K instruction data from ChatGPT~\cite{achiam2023gpt} to finetune LLaMA and obtain a strong instruction-following LLM. Vicuna~\cite{vicuna2023} adopted data from ShareGPT~\cite{sharegpt} to train LLaMA and achieved 90\% quality of ChatGPT. Orca~\cite{mukherjee2023orca} and Orca-2~\cite{mitra2023orca} adopt progressive learning and more data from GPT-4~\cite{achiam2023gpt} to further enhance the general abilities of LLMs. Different from the others, WizardLM~\cite{xu2024wizardlm} designed an evolve-instruct prompt to distill instruction data from ChatGPT with varying difficulty, encouraging LLMs to follow complex instructions. Zephyr~\cite{tunstall2023zephyr} employed DPO~\cite{rafailov2024direct} to align LLMs with human preference, while SPIN~\cite{chenself} devised a self-play method to achieve this.

For enhancing specific capabilities of LLMs, researchers used massive task-specific data to largely improve the desired abilities of LLMs such as math~\cite{luo2023wizardmath,tangmathscale}, coding~\cite{roziere2023code,guo2024deepseek}, and reasoning~\cite{lee2024llm2llm,ying2024llms}. Furthermore, some works developed specificalized LLMs in various domains such as medical~\cite{wang2023huatuo}, biology~\cite{labrak2024biomistral}, and chemistry~\cite{m2024augmenting}.

Our work aims to detect the knowledge deficiencies of LLMs and apply the appropriate remedies to repair these deficiencies, which can serve as a complement to or a patch for existing methods.

\subsection{Evaluation of LLMs}
Since LLMs have strong and broad capabilities, evaluating LLMs becomes a tough and widely concerned issue.

Some works constructed benchmarks or evaluation data to evaluate LLMs from general and specialized perspectives, such as natural language understanding~\cite{hendrycksmeasuring,li2023cmmlu,wang2024mmlu}, reasoning~\cite{frohberg2022crass,suzgun2023challenging,zhong2024agieval}, math~\cite{li2024gsm,liu2024mathbench}, coding~\cite{peng2024humaneval,zhang2024naturalcodebench}, etc.

Different from utilizing benchmarks with objective questions, some works started utilizing LLMs to evaluate LLMs subjectively. \citet{liu2023g} applied GPT-4 as an evaluator to assess the quality of generated text in various pesperctives~(such as coherence) and achieved better alignment with human evaluators. \citet{bai2024benchmarking} evaluated the performance of existing LLMs with self-evaluation and peer-evaluation, achieving more precise judgments of existing LLMs. Furthermore, \citet{zheng2024judging} devised a chat framework to evaluate LLMs based on the discussions among LLMs.

Different from previous works, our work utilizes relative entropy to diagnose the knowledge deficiencies in LLMs with the help of a knowledge base. We also follow previous work to adopt objective benchmarks to evaluate the performance of enhanced LLMs.

\begin{figure*}[t]
    \centering
    \includegraphics[width=1.0\linewidth]{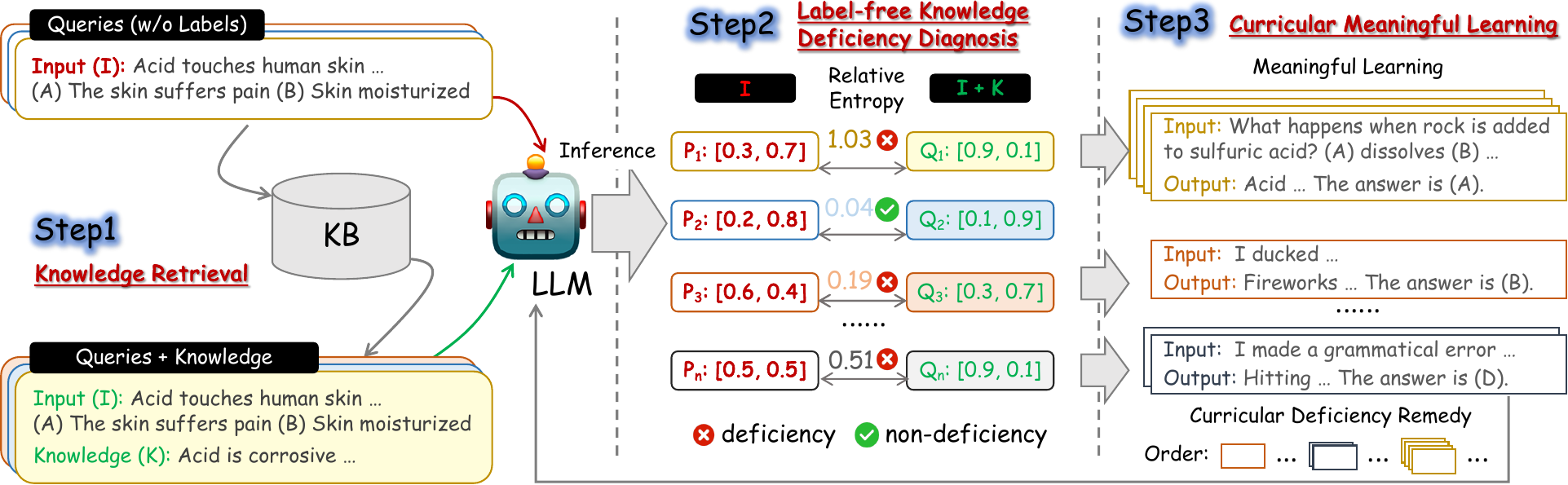}
    \caption{The whole workflow of LaMer aims to automatically diagnose and remedy the knowledge deficiencies in LLMs.}
    \Description{The framework of our proposed method.}
    \label{fig:method}
\end{figure*}

\section{Method}
We design a label-free curricular meaningful learning framework named LaMer, which utilizes user queries to efficiently diagnose and remedy the knowledge deficiencies in LLMs without labels. As illustrated in Figure~\ref{fig:method}, LaMer consists of 3 steps: (1)~\textbf{Knowledge Retrieval} retrieves relevant knowledge from an external knowledge base for each query to help diagnose knowledge deficiencies; (2)~\textbf{Label-free Knowledge Deficiency Diagnosis} leverages relative entropy to automatically diagnose and quantify the knowledge deficiencies in LLMs, which does not rely on labels; (3)~\textbf{Curricular Meaningful Learning} incorporates the idea of human conducting meaningful learning~\cite{tenenbaum2018building} to first adaptively synthesize examples in various scenarios for each knowledge deficiency, and then utilize a curricular deficiency remedy strategy to effectively and progressively repair the knowledge deficiencies from minor to severe.

\subsection{Knowledge Retrieval}
To introduce knowledge for the following deficiency diagnosis step, we employ an external knowledge base GenericsKB~\cite{bhakthavatsalam2020genericskb} to retrieve knowledge for the given query set $D$. GenericsKB is a large-scale resource containing 3.4M+ naturally occurring generic facts~(such as ``Trees remove carbon dioxide from the atmosphere'').

Specifically, to ensure the quality of GenericsKB, we first filter out facts with a confidence score lower than 0.7 and remove duplicates. Next, we employ FlagEmbedding~\cite{llm_embedder} to represent the facts in GenericsKB and each query in $D$ as dense embeddings. Finally, for each query $d\in D$, we apply cosine similarity to retrieve $m$ pieces of knowledge $K=\{k_1,\cdots,k_m\}$ from GenericsKB. The retrieved knowledge $K$ and example $d$ are used in step 2 to diagnose knowledge deficiencies of a specific LLM $\mathcal{L}$.

\subsection{Label-free Knowledge Deficiency Diagnosis}
\label{sec:re}
To diagnose the knowledge deficiencies of a specific LLM $\mathcal{L}$ in a label-free setting, we propose using relative entropy~\cite{kullback1951information}, also known as Kullback-Leibler divergence. This measure quantifies the additional information needed to transition from one distribution to another. Thus, by computing the relative entropy between predictive distributions of $\mathcal{L}$ before and after the introduction of knowledge, we can estimate the volume of information that this knowledge imparts to $\mathcal{L}$. If $\mathcal{L}$ exhibits a high relative entropy on this knowledge, it suggests that the model either lacks this knowledge or is unable to effectively integrate it into its problem-solving processes, a knowledge deficiency in $\mathcal{L}$ is diagnosed.

Specifically, given a query $d\in D$, which possesses a question $x$ and $n$ options $O = \{o_1,o_2,\cdots,o_n\}$. We first send $d$ to LLM $\mathcal{L}$ to obtain the negative log-likelihood of each option $o_i$ conditioned on $x$ and $O$. Hereafter, we acquire a prior distribution of $\mathcal{L}$ over $O$:
\begin{equation}
\label{eq:p}
\begin{aligned}
    p_i &= \mathcal{L}(y_i|x,O), \\
    P &= \text{Softmax}([p1, \cdots, p_n]),
\end{aligned}
\end{equation}
where $p_i$ denotes the negative log-likelihood of option $o_i$ conditioned on $x$ and $O$ according to $\mathcal{L}$. $P\in\mathbb{R}^n$ denotes the prior distribution of $\mathcal{L}$ over $O$. Softmax is the normalization function.

Secondly, for each retrieved knowledge $k\in K$ of $d$, we additionally introduce $k$ to $\mathcal{L}$ to fetch a knowledge-based posterior distribution of $\mathcal{L}$ over $O$:
\begin{equation}
\label{eq:q}
\begin{aligned}
    q_i &= \mathcal{L}(o_i|k,x,O), \\
    Q &= \text{Softmax}([q1, \cdots, q_n]),
\end{aligned}
\end{equation}
where $q_i$ denotes the negative log-likelihood of option $o_i$ after introducing $k$ to $\mathcal{L}$. $Q\in\mathbb{R}^n$ denotes the knowledge-based posterior distribution of $\mathcal{L}$ over $O$.

Subsquently, we compute the relative entropy $\text{RE}$ between $P$ and $Q$ to quantify the knowledge difficiency of LLM $\mathcal{L}$ on $k$:
\begin{equation}
\label{eq:re}
    \text{RE} = -\sum_{i=0}^mP_i\times (log(Q_i) - log(P_i)).
\end{equation}


Finally, after estimating RE on each knowledge of each query based on $\mathcal{L}$, we filter the knowledge and corresponding queries that result in an RE larger than a threshold $\tau$. Each filtered knowledge and its associated query are then treated as a unit, representing a knowledge deficiency in LLM $\mathcal{L}$. The RE is also a quantification of each knowledge deficiency in $\mathcal{L}$.

Note that there might be two situations: (1)~\textbf{Helpful}: the retrieved knowledge has a positive impact on $\mathcal{L}$, resulting in higher confidence in the correct option; (2)~\textbf{Misleading}: the retrieved knowledge has a negative impact on $\mathcal{L}$, resulting in higher confidence in the wrong option. We suppose that both situations can expose the knowledge deficiencies in $\mathcal{L}$. The first situation suggests $\mathcal{L}$ might not grasp this knowledge or cannot properly apply this knowledge to problem-solving, while the second situation indicates that $\mathcal{L}$ does understand this knowledge and is easily misled by it.

\begin{table}[t]
\centering
\caption{The divided groups of knowledge deficiencies and the corresponding size of synthesized examples.}
\begin{tabular}{lcc}
\toprule
\textbf{Group} & \textbf{RE Interval} & \textbf{Synthesized Examples} \\ \midrule
\textbf{Easy} & 0.1 $\le$ RE $<$ 0.4               & 1                    \\
\textbf{Normal} & 0.4 $\le$ RE $<$ 0.7                & 2                    \\
\textbf{Hard} & 0.7 $\le$ RE $<$ 1.0                & 3                    \\
\textbf{Unfair} & RE $\ge$ 1.0                & 4                    \\ \bottomrule
\end{tabular}
\label{tab:method}
\end{table}

\subsection{Curricular Meaningful Learning}
Humans adopt meaningful learning to induce and learn new knowledge through its application across diverse situations~\cite{tenenbaum2018building}, leading to a deep understanding and integration of knowledge. Moreover, Humans exploit curriculum learning~\cite{bengio2009curriculum} to effectively learn new knowledge by progressing from easy to hard levels. In light of this, we combine them and design curricular meaningful learning to effectively remedy the diagnosed knowledge deficiencies of $\mathcal{L}$.

Firstly, we employ a meaningful learning strategy to generate varying examples in diverse scenarios according to the severity of the knowledge deficiencies in $\mathcal{L}$. It is inspired by meaningful learning of humans~\cite{tenenbaum2018building} and the insights that LLMs typically require more tokens or examples to learn the knowledge if they have less prior understanding of it~\cite{ovadia2023fine,gekhman2024does}. This strategy can reduce the cost and make deficiency remedy more efficient. Specifically, for the diagnosed knowledge deficiencies of $\mathcal{L}$, we divide them into 4 groups according to the severity~(relative entropy) of them. For each group, we heuristically assign a number, which indicates the number of diverse examples we should synthesize for each knowledge deficiency in the group. The detailed groups and assigned numbers are illustrated in Table~\ref{tab:method}. Subsequently, we adopt a strong LLM ChatGPT~\cite{achiam2023gpt} to synthesize the specified number of examples for the knowledge deficiencies in each group. The knowledge deficiency~(knowledge and the corresponding query) is harnessed to guide the data synthesis process. Each synthesized example contains an input $X$ as the query and an output $Y$ as the response.

Secondly, we devise a curricular deficiency remedy strategy to progressively remedy the knowledge deficiencies in LLM $\mathcal{L}$ from minor to severe. Specifically, we sort the generated examples in ascending order based on the severity of their corresponding knowledge deficiencies, and then feed them into training $\mathcal{L}$ sequentially. For each generated example $<X, Y>$, the training process is in an autoregressive manner to maximize a conditional probability:
\begin{equation}
    \mathcal{L}(X,Y,\theta)=-\sum_t{log_{p_\theta}(Y_t|X,Y_{<t})},
\end{equation}
where $\theta$ denotes parameters of $\mathcal{L}$. This approach results in an updated $\mathcal{L}$ with its deficiencies effectively remedied.

\begin{table*}[t]
\centering
\tabcolsep=6.5pt
\caption{Overall performance of LaMer and baselines. \textbf{Bold} numbers denote the best performance among all methods. \textbf{\textit{Average}} denotes the performance averaged across all benchmarks.}
\begin{tabular}{lllcccccccc}
\toprule
\textbf{LLMs} & \textbf{Size} & \textbf{Methods} & \textbf{Comm.} & \textbf{AGIEval} & \textbf{ARC} & \textbf{MMLU} & \textbf{BBH} & \textbf{CRASS} & \textbf{GSM-Plus} & \textit{\textbf{Average}} \\ \midrule
 & & Base & 67.56 & 32.82 & 74.15 & 49.75 & 28.47 & 71.67 & 29.91 & 50.62 \\
 & & AugGPT  & 74.53 & 33.03 & 76.93 & 52.56 & \textbf{33.82} & 81.67 & 13.78 & 52.33 \\
 & & Naive  & 67.42 & 33.60 & 72.66 & 50.58 & 32.22 & 80.00 & \textbf{31.19} & 52.52 \\
 & & Single  & 69.47 & \textbf{34.87} & 74.85 & 51.44 & 29.53 & 80.00 & 29.03 & 52.74 \\
\multirow{-5}{*}{\textbf{Mistral}} & \multirow{-5}{*}{7B} & LaMer (Ours) & \textbf{75.10} & 34.50 & \textbf{77.22} & \textbf{54.52} & 33.72 & \textbf{88.33} & 29.23 & \textbf{56.09} \\ \midrule
 & & Base & 74.84 & 39.45 & 86.29 & 59.72 & 39.85 & 76.67 & \textbf{61.93} & 62.82 \\
 & & AugGPT  & 76.75 & 38.13 & 86.04 & \textbf{60.29} & 36.60 & 76.67 & 14.92 & 55.63 \\
 & & Naive  & 71.28 & 33.86 & 84.50 & 57.86 & \textbf{40.93} & 83.33 & 61.56 & 61.90 \\
 & & Single & 76.36 & 39.72 & 84.72 & 58.97 & 36.53 & 76.67 & 55.85 & 61.26 \\
\multirow{-5}{*}{\textbf{LLaMA-3}} & \multirow{-5}{*}{8B} & LaMer (Ours)  & \textbf{78.15} & \textbf{40.85} & \textbf{86.41} & 60.24 & 39.69 & \textbf{88.33} & 59.22 & \textbf{64.70} \\ \midrule
 & & Base & 71.88 & 42.10 & 85.48 & 59.53 & 37.58 & 85.00 & 61.79 & 63.34 \\
 & & AugGPT & \textbf{79.41} & 42.71 & \textbf{89.78} & \textbf{63.66} & 40.00 & \textbf{90.00} & 14.75 & 60.04 \\
 & & Naive & 75.62 & 42.80 & 88.40 & 62.50 & 39.34 & 86.67 & \textbf{61.82} & 65.31 \\
 & & Single & 76.41 & 44.30 & 87.64 & 61.98 & 39.69 & 80.00 & 56.62 & 63.81 \\
\multirow{-5}{*}{\textbf{Qwen2}} & \multirow{-5}{*}{7B} & LaMer (Ours) & 78.13 & \textbf{45.02} & 88.62 & 62.48 & \textbf{40.56} & 86.67 & 61.44 & \textbf{66.13} \\ \midrule
 & & Base & 53.26 & 25.80 & \textbf{49.97} & 33.73 & 23.29 & 38.33 & \textbf{7.08} & 33.07 \\
 & & AugGPT & 55.42 & 25.76 & 47.73 & 33.69 & 24.31 & 36.67 & 3.95 & 32.50 \\
 & & Naive & 55.38 & 25.78 & 47.79 & 32.23 & 24.24 & 36.67 & 5.99 & 32.58 \\
 & & Single & 52.85 & 24.51 & 46.67 & 34.00 & 25.03 & 31.67 & 5.86 & 31.51 \\
\multirow{-5}{*}{\textbf{Gemma-1.1}} & \multirow{-5}{*}{2B} & LaMer (Ours) & \textbf{55.81} & \textbf{25.81} & 48.91 & \textbf{34.40} & \textbf{25.26} & \textbf{41.67} & 6.69 & \textbf{34.08} \\ \bottomrule
\end{tabular}
\label{tab:results}
\end{table*}

\section{Experiments}
\subsection{Investigated LLMs}
\label{sec:llms}
We employ 4 different open-source LLMs for experiments to illustrate the general applicability of our proposed LaMer: 
\begin{itemize}
    \item \textbf{Mistral-7B-Instruct-V0.2}~\cite{jiang2023mistral}~(denoted as Mistral) is an efficient instruction-tunned LLM developed by Mistral AI.
    \item \textbf{LLaMA-3-8B-Instruct}~\cite{llama3modelcard}~(denoted as LLaMA-3) is a dense LLM with massive pre-training on extremely large corpora, which is developed by Meta.
    \item \textbf{Qwen2-7B-Instruct}~\cite{qwen2}~(denoted as Qwen2) is a powerful multilingual LLM developed by Alibaba Cloud.
    \item \textbf{Gemma-1.1-2B-IT}~\cite{team2024gemma}~(denoted as Gemma-1.1) is an instruction-tunned LLM developed by DeepMind.
\end{itemize}

\subsection{Baselines}
We adopt a wide range of baselines for comprehensive comparisons:
\begin{itemize}
    \item \textbf{Base} employs the base LLMs in Section~\ref{sec:llms} to answer questions in each benchmark.
    \item \textbf{AugGPT}~\cite{dai2023auggpt} uses ChatGPT~\cite{achiam2023gpt} to generate questions and answers to augment LLMs with SFT. AugGPT does not provide chain-of-thought~\cite{wei2022chain} in generated examples.
    \item \textbf{Naive} is an SFT method that randomly samples several pieces of knowledge from the knowledge base to synthesize new examples without considering whether the specific LLM possesses deficiencies on the sampled knowledge.
    \item \textbf{Single} follows a similar process to LaMer but synthesizes only 1 example per knowledge deficiency. Hence, the training data of Single is 40\% of LaMer or the other methods.
\end{itemize}

\subsection{Evaluation Benchmarks}
We choose 7 OOD benchmarks ranging from reasoning to language understanding, to evaluate the performance of LaMer and baselines:
\begin{itemize}
    \item \textbf{Comm.}~\cite{xiong2023examining} is a collection of 6 commonsense reasoning datasets: An abductive reasoning dataset $\alpha$NLI~\cite{bhagavatulaabductive}, a commonsense reasoning dataset CommonsenseQA~\cite{talmor2019commonsenseqa}, a commonsense causal reasoning dataset COPA~\cite{roemmele2011choice}, a social interaction reasoning dataset SocialIQa~\cite{sap2019social}, a physical interaction reasoning dataset PIQA~\cite{bisk2020piqa}, and an implicit strategy reasoning dataset StrategyQA~\cite{geva2021did}.
    \item \textbf{AGIEval}~\cite{zhong2024agieval} consists of diverse sets of standardized tests ranging from college admission tests~(such as GRE and GMAT) to national civil service examinations.
    \item \textbf{ARC}~\cite{clark2018think} is the AI2 Reasoning Challenge, which is a benchmark of science exams spanning Grade 3 to Grade 9 with easy~(ARC-e) and challenge~(ARC-c) subsets.
    \item \textbf{MMLU}~\cite{hendrycksmeasuring} aims to evaluate language comprehension, knowledge, and reasoning skills of LLMs with 57 tasks.
    \item \textbf{BBH}~\cite{suzgun2023challenging} is a subset of Big-Bench~\cite{srivastava2023beyond}, which contains 23 hardest tasks focusing on challenging scenarios.
    \item \textbf{CRASS}~\cite{frohberg2022crass} measures the counterfactual reasoning abilities of language models.
    \item \textbf{GSM-Plus}~\cite{li2024gsm} is a comprehensive math benchmark for evaluating the robustness of LLMs. We only keep the examples that possess valid answers for evaluation.
\end{itemize}

\subsection{Implementation Details}
For the knowledge retrieval step, we choose e-CARE~\cite{du2022care} and GSM8K~\cite{cobbe2021training}, discarding the labels to obtain query set $D$. The size of GenericsKB after filtering is 200K. We utilize FlagEmbedding~\cite{llm_embedder} with beg-large-en-v1.5 to encode facts and queries. We retrieve $m=4$ facts as the knowledge for each query. Since GSM8K lacks options and is far from GenericsKB, we choose ChatGPT to generate distractors and $m=4$ pieces of knowledge for GSM8K.

For the label-free knowledge deficiencies diagnosis step, the knowledge and corresponding query with an RE higher than $\tau=0.1$ is treated as a knowledge deficiency of LLM $\mathcal{L}$. The size of selected knowledge deficiencies can refer to Table~\ref{tab:difi} in Appendix~\ref{app:difi}.

For the meaningful learning strategy in curricular meaningful learning, we utilize a carefully designed prompt to instruct ChatGPT to synthesize examples. Finally, we synthesize 3,750 examples to enhance Mistral, Qwen2, and Gemma-1.1, while 1,250 examples are synthesized to enhance LLaMA-3 due to denser knowledge in it.

For the curricular deficiency remedy strategy, we adopt LoRA~\cite{hulora} for parameter-efficient fine-tuning. The rank $r$ and $\alpha$ of LoRA are 128 and 8, respectively. We train LaMer for 3 epochs with a learning rate of $5e$-$5$. The batch size is 32. The optimizer we used is Adam~\cite{kingma2014adam}. Two NVIDIA A100 80GB PCIe GPUs are used for training and the following evaluation.

For AugGPT and Naive, we randomly sample 3,750 facts from GenericsKB and the generated knowledge of GSM8K to generate the same number of training examples as LaMer for each LLM. While for Single, we randomly sample one example for each knowledge deficiency from the training data of LaMer. Therefore, Single enhances Mistral, Qwen2, and Gemma-1.1 with 1,500 examples, and it utilizes 600 examples to enhance LLaMA-3. The whole data synthesis process is the same as LaMer and the training setup is also the same as LaMer. All prompts for the process of GSM8K and data synthesis can refer to Appendix~\ref{app:prompts}. We utilize gpt-3.5-turbo-0125 for all relevant implementations based on ChatGPT.

\subsection{Evaluation Details}
For all benchmarks, we utilize free-form generation to evaluate all methods, the evaluation prompts for different LLMs can refer to the Appendix~\ref{app:prompts_eval}. The performance of each method on each benchmark is averaged across all tasks in the corresponding benchmark.

\subsection{Overall Results}
The overall results of LaMer and baselines are shown in Table~\ref{tab:results}, from which we can have the following observations:

(1)~On average, LaMer can outperform all baseline methods across different base LLMs, which is mainly due to the effectiveness of LaMer in diagnosing and remedying knowledge deficiencies without labels. This also reveals the general applicability of LaMer, making it a plug-and-play method to improve LLMs.


(2)~All data augmentation methods can surpass the base LLM on most benchmarks, while Nive and Single obtain performance drops than base LLaMA-3 and base Gemma-1.1, this is mainly because LLaMA-3 and Gemma-1.1 possess dense knowledge in its parameters. Naive and Single could supplement some knowledge to them but cause them to forget more knowledge.

(3)~Interestingly, Single, which is trained with 40\% training data of LaMer~(one example for each knowledge deficiency), can achieve comparable performance with Naive on Mistral and LLaMA-3 with much fewer training data. This indicates Naive method can produce more redundant data that the base LLM already possesses, while detecting knowledge deficiencies in LLMs can apply targeted improvements, making it more efficient and less costly than baselines.

(4)~LaMer excels Single across all base LLMs, this is because more severe knowledge deficiencies require more and diverse examples for LLMs to effectively remedy them~\cite{gekhman2024does}. LaMer synthesizes more examples for knowledge deficiencies with higher severity, whereas Single generates only one example for each knowledge deficiency. As a result, many knowledge deficiencies in the LLMs are not adequately remedied by Single.

(5)~AugGPT achieves the worst results on GSM-Plus benchmark, this is because math problems demand multiple reasoning steps to solve. AugGPT directly offers the answer to the math problems, making it hard to develop precise calculations to arrive at the right answer. Furthermore, AugGPT can reach the best performance on several benchmarks~(such as ARC on Qwen2), this is because problems in these benchmarks might be better answered with just a step, which is also indicated in \citet{zhangautomatic}.

(6)~Qwen2 could exceed LLaMA-3 across different methods with fewer parameters, we suppose this is mainly due to the massive post-training efforts on Qwen2.

\section{Case Study}
We provide a case to demonstrate how LaMer diagnoses and remedies a knowledge deficiency of Mistral in Figure~\ref{fig:case}:

\begin{figure}[t]
    \centering
    \includegraphics[width=0.9\linewidth]{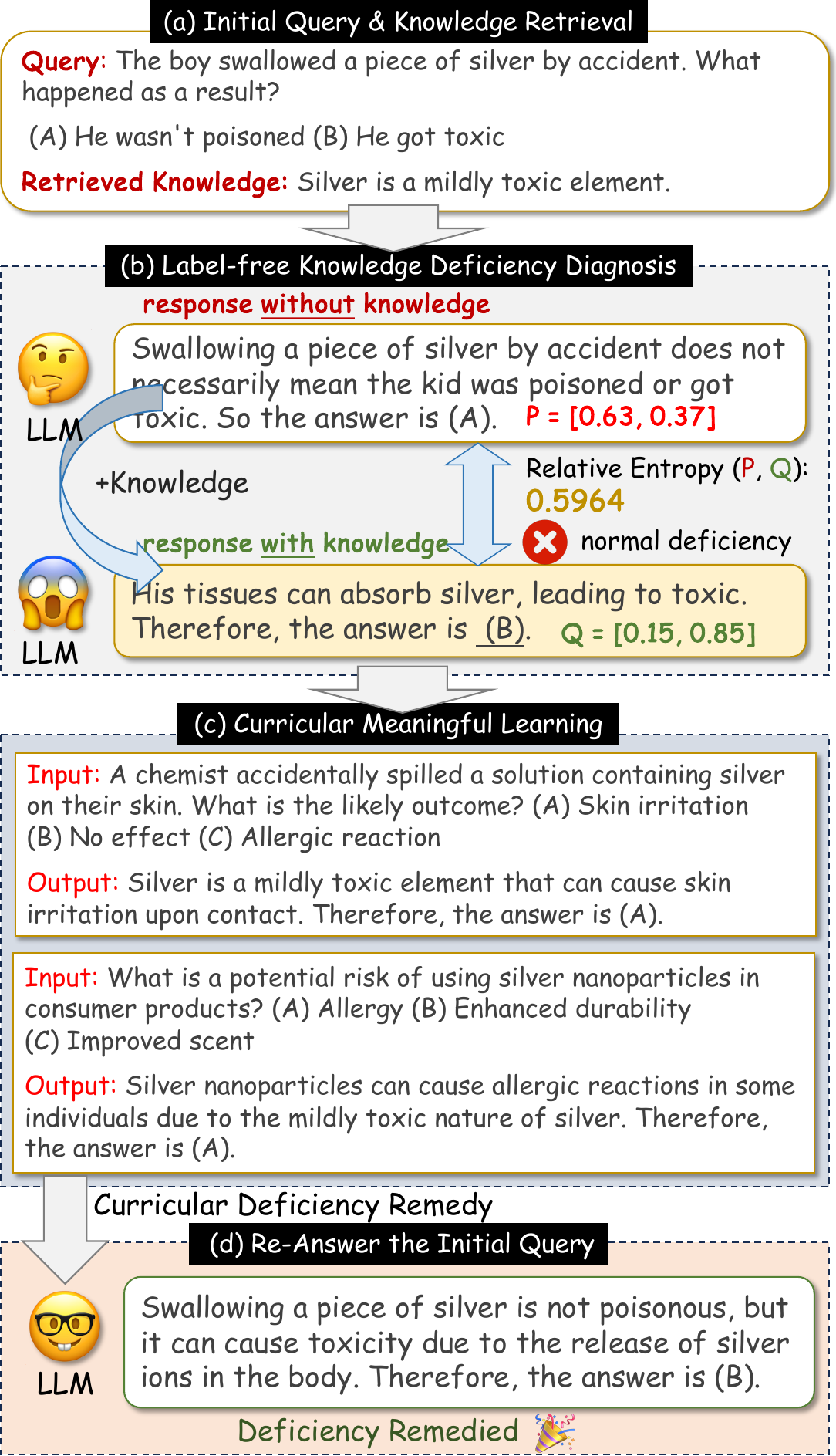}
    \caption{A case showing how LaMer diagnoses and remedies a knowledge deficiency in Mistral.}
    \label{fig:case}
    \Description{A case of LaMer on Mistral}
\end{figure}

(a)~For an initial query~(without label) about ``silver'', we can retrieve the knowledge that ``silver is a mildly toxic element''.

(b)~The LLM offer a wrong response which produce a distribution $P=[0.63, 0.37]$ over the options. After providing the LLM with the retrieved knowledge, the LLM offers a new response with a different distribution $Q=[0.15, 0.85]$. The relative entropy between $P$ and $Q$ is 0.5964, which means the knowledge brings a lot of information to the LLM. Hence, the combination of the knowledge and the query is a knowledge deficiency of the LLM.

(c)~According to Table~\ref{tab:method}, this knowledge deficiency is in normal group. Thus, we ask ChatGPT to synthesize two examples of this deficiency. These two examples are two applications in two different scenarios of the knowledge in this deficiency. Next, the synthesized examples will be used to train the LLM with other synthesized data.

(d)~After curricular meaningful learning, Mistral offers the correct answer to the initial query. The deficiency is remedied.

\section{Further Analysis}
To further investigate the strengths and effectiveness of LaMer, we design several ablation studies and in-depth analyses: (1)~comparisons between LaMer and label-reliant methods to demonstrate the efficiency and strengths of LaMer; (2)~an effectiveness analysis of LaMer on remedying deficiencies by obtain the statistics on remedied examples of each method; (3)~an ablation study to investigate the effect of curricular deficiency remedy strategy; (4)~a visualization analysis on the distribution of data synthesized by various methods to explore potential causes for the strengths of LaMer; (5)~a significance analysis to inspect the roles of helpful and misleading situations in label-free knowledge deficiency diagnosis.

\begin{table}[t]
\caption{Effectiveness of different methods on detecting difficiencies of Mistral~\cite{jiang2023mistral} based on e-CARE~\cite{du2022care}}
\centering
\tabcolsep=5pt
\begin{tabular}{lcccc}
\toprule
\textbf{Method}  & \textbf{Label-free} & \textbf{Precision} & \textbf{Recall} & \textbf{F1}    \\ \midrule
Golden Label     & \cellcolor[HTML]{FFCCC9}No       & 100       & 100    & 100   \\
Perplexity       & \cellcolor[HTML]{FFCCC9}No       & 48.46     & 34.24  & 40.10 \\
Random           & \cellcolor[HTML]{9AFF99}Yes        & 35.23     & 35.23  & 35.23 \\
Relative Entropy & \cellcolor[HTML]{9AFF99}Yes        & 40.34     & 64.30  & 49.58 \\ \bottomrule
\end{tabular}
\label{tab:preliminary}
\end{table}

\begin{figure}[t]
    \centering
    \includegraphics[width=0.9\linewidth]{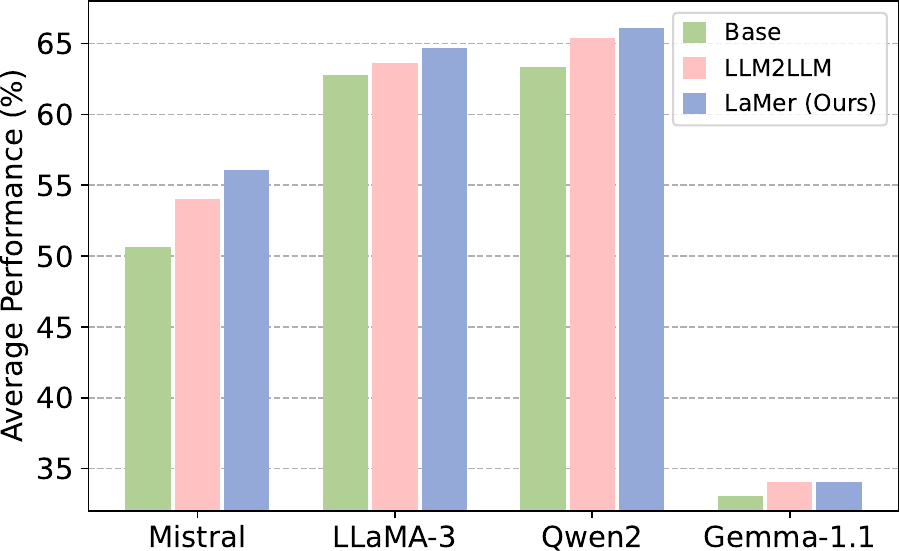}
    \caption{The average performance across all evaluation benchmarks of base LLMs, LLM2LLM, and LaMer.}
    \label{fig:llm2llm}
    \Description{the performance of base LLM, LLM2LLM and LaMer}
\end{figure}

\subsection{Comparisons to Label-reliant Methods}
We conduct an initial analysis to reveal the efficiency of different methods in diagnosing deficiencies based on Mistral-7B-Instruct-v0.2~\cite{jiang2023mistral} and e-CARE~\cite{du2022care}. Implementation details can refer to Appendix~\ref{app:preliminary}. The results are shown in Table~\ref{tab:preliminary}, we can have the following observations: our proposed relative entropy method significantly outperforms the perplexity method by recalling more deficiencies for the specific LLM. This proves the advantage and feasibility of our proposed relative entropy method.

Furthermore, we adopt a label-reliant method LLM2LLM for comparison. LLM2LLM~\cite{lee2024llm2llm} utilizes labeled data to identify erroneous examples in existing datasets, and then synthesizes similar examples to improve a specific LLM. We use the labels of e-CARE and GSM8K to obtain 3,750 error examples, and then we generated similar examples based on the error examples. The number of training examples is the same as LaMer for each LLM. The results are shown in Figure~\ref{fig:llm2llm}~(full results can refer to Appendix~\ref{app:analysis}), we can find that: LaMer has varying advantages over LLM2LLM on different base LLMs. Although LLM2LLM can precisely detect the error examples of each LLM, the error examples are limited to the given initial dataset. LaMer diagnoses knowledge deficiencies, which can improve the coverage of diagnosed deficiencies in each LLM.

\begin{figure}[t]
    \centering
    \includegraphics[width=\linewidth]{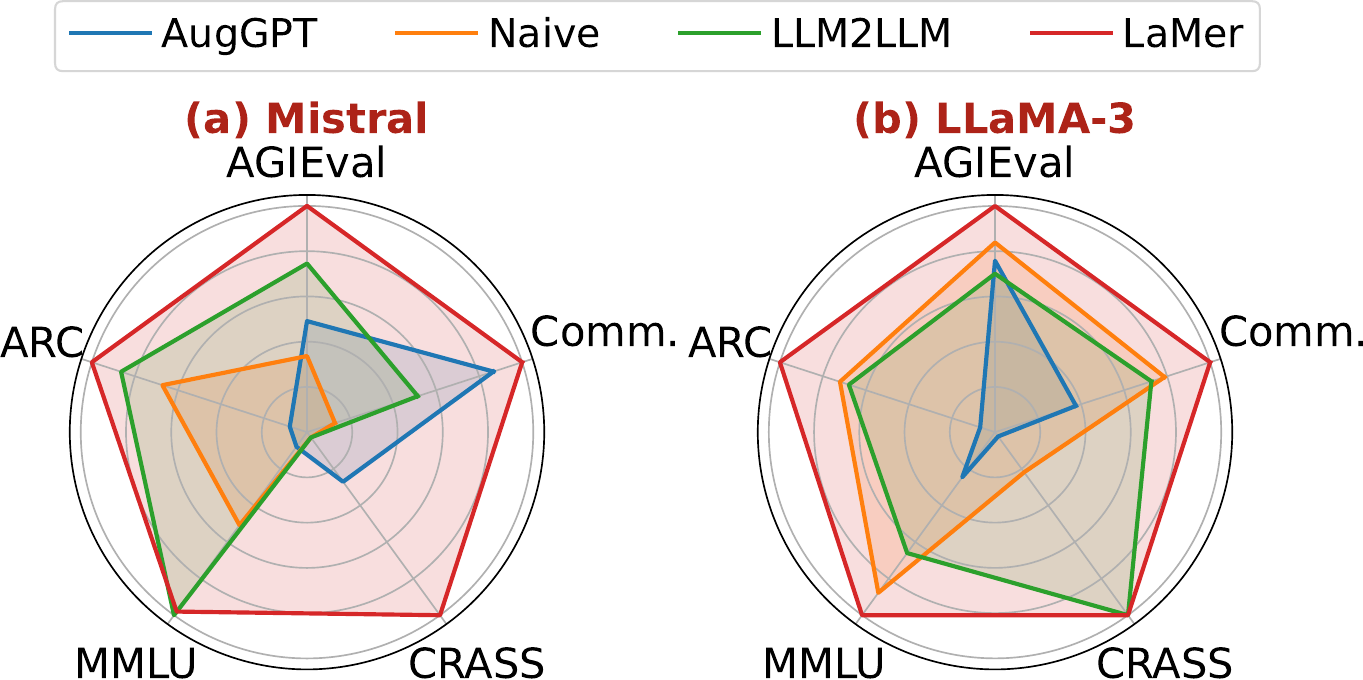}
    \caption{Normalized visualization of remedied examples on (a) Mistral and (b) LLaMA-3 across different benchmarks.}
    \label{fig:analysis3}
    \Description{Visualization remedied examples on each benchmark.}
\end{figure}

\subsection{Effectiveness in Remedying Decificiencies}
We conduct an analysis to examine if LaMer can effectively remedy more error examples. Specifically for each method, we first tally the number of examples in each benchmark that base LLM is wrong but correct after the enhancement by each method~(denote as remedied examples). After that, we normalize the values across different methods. Finally, we use spider charts to visualize them. We choose Mistral and LLaMA-3 for investigation. The results are shown in Figure~\ref{fig:analysis3}, we can infer:

(1)~LaMer has advantages over the baselines on all benchmarks, this is attributed to that LaMer can diagnose more deficiencies via relative entropy, and remedy them more efficiently via curricular meaningful learning.




(2)~In Table~\ref{tab:results}, although AugGPT can exceed LaMer on some benchmarks, it does not get the upper hand in Figure~\ref{fig:analysis3}. This is mainly due to AugGPT cannot diagnosis the deficiencies of LLMs, and encouraging LLMs to give answers directly is not effective in remedying the deficiencies.


\begin{table*}[t]
\centering
\tabcolsep=7pt
\caption{Ablation study results of the effect of the curricular deficiency remedy, LaMer$^*$ denotes that we train corresponding LLMs with randomly shuffled data. Bold numbers denote better performance between LaMer and LaMer$^*$.}
\begin{tabular}{lllcccccccc}
\toprule
\textbf{LLMs} & \textbf{Size} & \textbf{Methods} & \textbf{Comm.} & \textbf{AGIEval} & \textbf{ARC} & \textbf{MMLU} & \textbf{BBH} & \textbf{CRASS} & \textbf{GSM-Plus} & \textit{\textbf{Average}} \\ \midrule
\multirow{2}{*}{\textbf{Mistral}} & \multirow{2}{*}{7B} & LaMer & \textbf{75.10} & \textbf{34.50} & 77.22 & \textbf{54.52} & \textbf{33.72} & \textbf{88.33} & \textbf{29.23} & \textbf{56.09} \\
 & & LaMer$^*$ & 74.90 & 34.07 & \textbf{77.48} & 54.30 & 33.66 & 86.67 & 28.42 & 55.64 \\ \midrule
\multirow{2}{*}{\textbf{LLaMA-3}} & \multirow{2}{*}{8B} & LaMer & \textbf{78.15} & 40.85 & \textbf{86.41} & \textbf{60.24} & \textbf{39.69} & \textbf{88.33} & 59.22 & \textbf{64.70} \\
 & & LaMer$^*$ & 77.67 & \textbf{40.93} & 86.28 & 60.07 & 38.47 & \textbf{88.33} & \textbf{59.55} & 64.47 \\ \midrule
\multirow{2}{*}{\textbf{Qwen2}} & \multirow{2}{*}{7B} & LaMer & \textbf{78.13} & \textbf{45.02} & \textbf{88.62} & \textbf{62.48} & \textbf{40.56} & \textbf{86.67} & \textbf{61.44} & \textbf{66.13} \\
 & & LaMer$^*$ & 77.51 & 44.33 & 88.36 & 62.12 & 39.80 & 85.00 & 61.36 & 65.50 \\ \midrule
\multirow{2}{*}{\textbf{Gemma-1.1}} & \multirow{2}{*}{2B} & LaMer & \textbf{55.81} & \textbf{25.81} & \textbf{48.91} & \textbf{34.40} & \textbf{25.26} & \textbf{41.67} & \textbf{6.69} & \textbf{34.08} \\
 & & LaMer$^*$ & 55.05 & 25.32 & 48.63 & 34.16 & 25.18 & \textbf{41.67} & 6.60 & 33.80 \\ \bottomrule
\end{tabular}

\label{tab:ablation}
\end{table*}

\subsection{Effect of Curricular Deficiency Remedy}
To investigate the effect of curricular deficiency remedy in LaMer, we randomly shuffle the data synthesized by LaMer to train the base LLMs. All the training and evaluation settings are the same as LaMer. We denote LaMer with shuffled data as LaMer$^*$. The overall results are shown in Table~\ref{tab:ablation}, we can observe that:

(1)~LaMer can outperform LaMer$^*$ on almost all benchmarks regarding each LLM. This is mainly because remedying less severe deficiencies helps remedy more severe ones. Curricular meaningful learning can make LLMs learn new knowledge more efficiently, just like humans conducting knowledge learning~\cite{cui2023efficient}.

(2)~When switched to randomly shuffled training data, LaMer$^*$ only suffers relatively small performance drops on each base LLM. This claims that the performance of LaMer improvement is stable and LaMer has robust applicability. 

(3)~LaMer$^*$ could achieve higher performance than the baselines in Table~\ref{tab:results} on Mistral, LLaMA-3, and Qwen2. It clarifies that the advantages of LaMer are not solely due to the curricular deficiency remedy strategy, but primarily stem from the deficiencies diagnosis process based on relative entropy. This highlights the merits of our relative-entropy-based deficiencies diagnosis process.

\begin{figure}[t]
    \centering
    \includegraphics[width=0.85\linewidth]{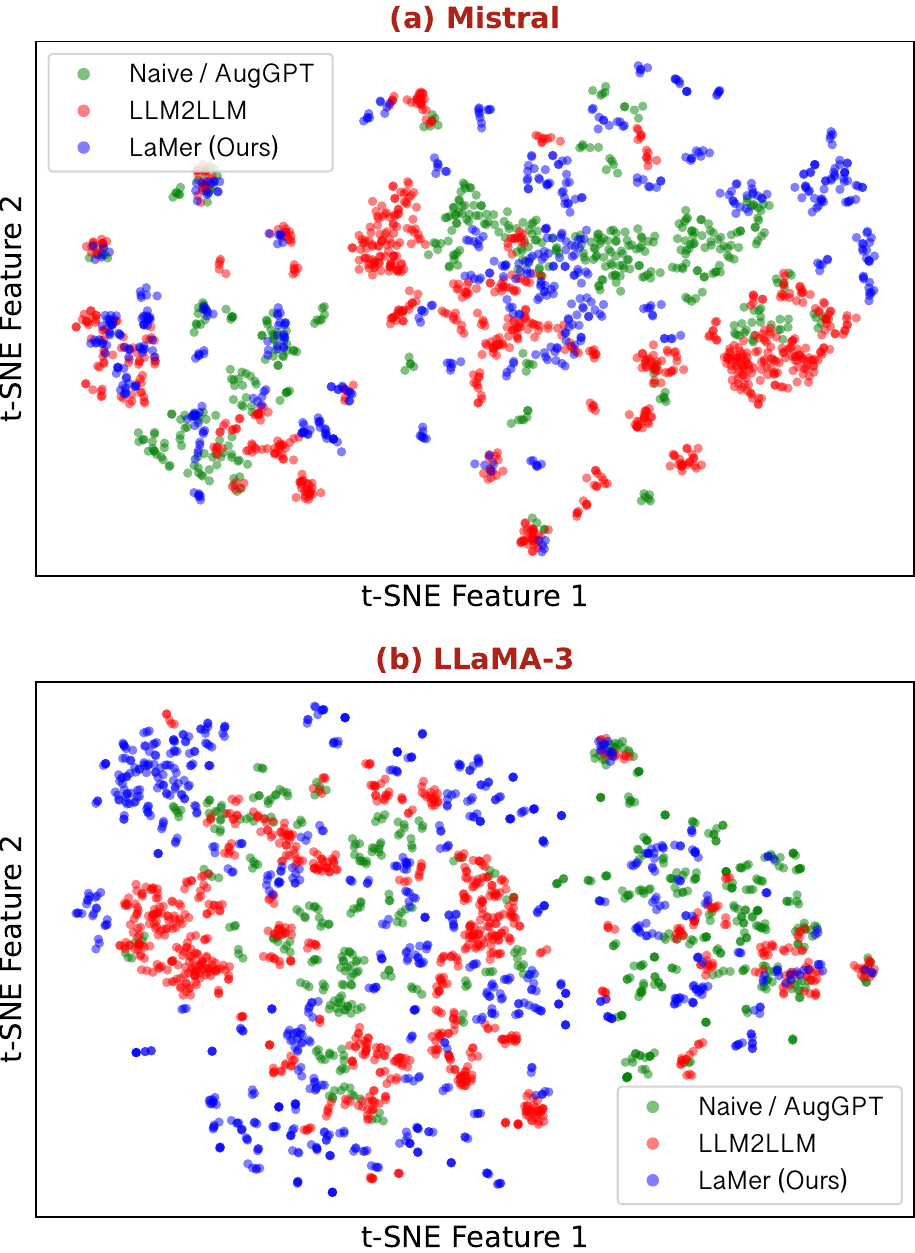}
    \caption{The distribution of synthesized data by baselines and LaMer on (a)~Mistral and (b)~LLaMA-3.}
    \label{fig:analysis2}
    \Description{}
\end{figure}

\subsection{Causes for the Advantages of LaMer}
To explore the potential causes for the advantages of LaMer, we visualize the data synthesized by baselines and LaMer into 2D space. Specifically, we utilize FlagEmbedding~\cite{llm_embedder} to represent each example in the synthesized data, and then reduce the dimensionality to 2 with the help of t-SNE~\cite{van2008visualizing}. Finally, we employ DBSCAN~\cite{ester1996density} to discover clusters and remove the noise data points. The eps and min samples of DBSCAN are 1.5 and 3, respectively. We also adopt Mistral and LLaMA-3 for experiments. The visualization is shown in Figure~\ref{fig:analysis2}, from which we can infer that, data synthesized by LaMer has a significantly higher proportion at the outer edges of the whole distribution~(LaMer and baselines). This might be a potential cause for the advancement of LaMer, since LaMer can utilize relative entropy to effectively discover more deficiencies.


\begin{figure}[t]
    \centering
    \includegraphics[width=0.85\linewidth]{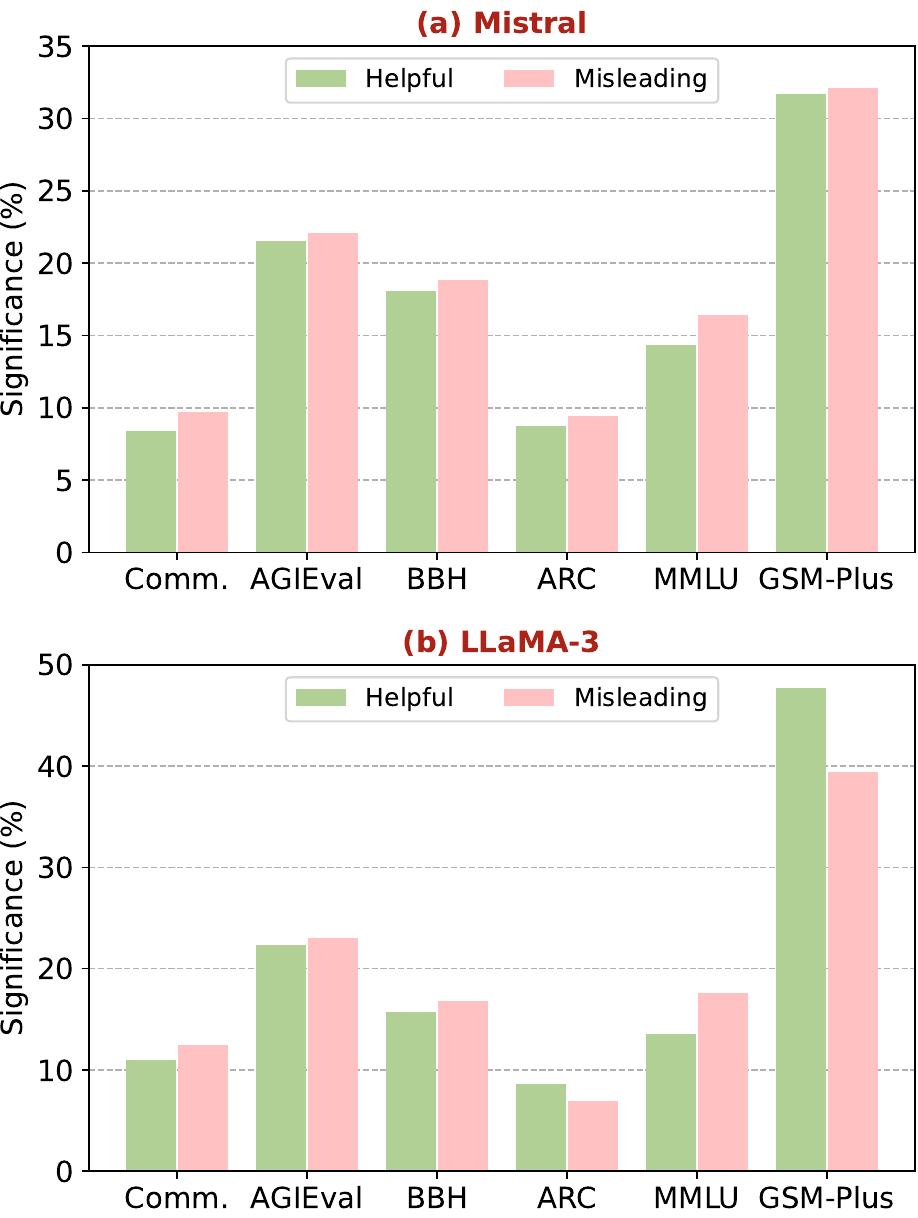}
    \caption{The portion of unique examples that are remedied based on the knowledge deficiencies caused by helpful and misleading knowledge across all evaluation benchmarks on (a)~Mistral and (b)~LLaMA-3.}
    \label{fig:analysis1}
    \Description{The significance of deficiencies caused by helpful and misleading knowledge.}
\end{figure}

\subsection{Significance of Deficiencies Caused by Helpful and Misleading Knowledge}
When diagnosing knowledge deficiencies, they can be caused by two kinds of knowledge: (1)~helpful knowledge which has a very positive impact on the correct answer; (2)~misleading knowledge which could lead LLMs to choose the wrong answer with much higher confidence than the right answer.

To find out the significance of the deficiencies caused by these two kinds of knowledge, we split the discovered knowledge deficiencies into two groups~(Helpful and Misleading) with the help of golden labels. Then we respectively train the LLM with data synthesized based on Helpful and Misleading deficiencies. Finally, after the evaluation process, we respectively analyze the unique remedied examples brought by the data synthesized based on Helpful and Misleading deficiencies. The results are shown in Figure~\ref{fig:analysis1}, from which we can have the following conclusions:

(1)~Deficiencies caused by helpful and misleading knowledge have similar and undeniable significance, which means helpful and misleading knowledge can assist LLMs in remedying similar numbers but different examples. This is due to that both kinds of knowledge can indicate valid knowledge deficiencies of corresponding LLMs to effectively boost them.

(2)~Deficiencies caused by misleading knowledge have slightly greater significance than those caused by helpful knowledge. This is mainly because deficiencies indicated by misleading knowledge often have a larger impact, and misleading knowledge is often more difficult and complex than helpful knowledge.

(3)~On the benchmarks where all methods have poorer performance in Table~\ref{tab:results}, deficiencies caused by both kinds of knowledge have larger significance. Owing to the LLMs have more pronounced knowledge deficiencies on benchmarks with poorer performance.

\section{Conclusion}
In this paper, we design a label-free curricular meaningful learning framework~(LaMer) based on relative entropy to first automatically discover the knowledge deficiencies from massive label-free user queries. Then we devise curricular meaningful learning which consists of a meaningful learning strategy and a curricular deficiency remedy strategy, to efficiently and effectively remedy the discovered knowledge deficiencies of corresponding LLM. The experiments show that our proposed LaMer can improve the coverage of diagnosed deficiencies, and surpass the baselines, making LLMs enhancement free of labeled data. The relative-entropy-based deficiency method provides a robust, efficient, and label-free deficiency diagnostic tool for existing LLMs to further unlock their potential.

\bibliographystyle{ACM-Reference-Format}
\bibliography{sample-base}

\appendix
\section{Prompts for Implementation}
\label{app:prompts}
\subsection{Prompt for Process GSM8K}

\begin{tcolorbox}[title = {Prompt for Processing GSM8K}, left = 1mm, right = 1mm, top = 1mm, bottom = 1mm, boxsep = 0mm]
You are an expert in math problems. 
\\ \\
First, please generate some background knowledge which can be used to solve this question. The knowledge SHOULD be general and applicable, Your generated knowledge CANNOT be question-specific. Like commonsense reasoning, I can give the knowledge "Acid is corrosive." Just list the knowledge. 
\\ \\
Second, please answer this question by giving a short explanation and then give the answer.
Third, please generate some distractors for your answer and index them like ``(A)\_\_ (B)\_\_ ···''.
\\ \\
The answer and distractors should be as concise as possible. Your response SHOULD follow the following format:
Background Knowledge: [The generated knowledge]
Explanation: [Steps to achieve the answer]
Answer: [A pure math part]
Distractors: [Wrong answers]
\end{tcolorbox}

Since GSM8K~\cite{cobbe2021training} contains math problems and without options, we adopt ChatGPT to generate distract options and knowledge for each query in GSM8K. The following is the prompt:

\subsection{Prompt for Synthesizing Data}
\begin{tcolorbox}[title = {LaMer: e-CARE}, left = 1mm, right = 1mm, top = 1mm, bottom = 1mm, boxsep = 0mm]
You are an expert in creating reasoning and language understanding questions.
\\ \\
Here are the requirements:\\
1. You are given a fact and some examples for reference, the fact can implicitly guide the solution for the reference examples. You should understand the Internal Mechanism of this guidance to make new examples.\\
2. Each created example should contain a question, several options (>=3), an answer, and a short explanation for the answer.\\
3. The options should be a short sentence or phrase rather than a single word whenever possible.\\
4. Don't make any commonsense mistakes and ensure that your solutions are accurate.\\
5. You SHOULD propose totally new reasoning questions in various areas, including but not limited to history, law, medical, math, science, computer science, psychology, AI, politics, economics, etc. \\
6. NOTE that the fact should be an implicit explanation for obtaining the true answer, which means the fact SHOULD NOT appear explicitly in the questions or the options. \\
7. Only one option is the correct answer, the other options should be much less plausible than the correct option or they are just wrong options. \\
8. Therefore is no explanation in the reference examples, you SHOULD generate an explanation first and then give the answer for your generated new questions. \\
9. The question could be in any form, such as "Why, What, How, Which" etc. You can also add a premise to form a question. \\
10. The created examples cannot be different just in nouns.
\\\\
Reference: \\
Knowledge: \{\quad\quad\} \\
Examples: \{\quad\quad\}
\\\\
You MUST generate \{\quad\quad\} new examples. The examples MUST be totally different from each other and the reference examples. Please return your response in the form:\\
Question: [QUESTION]\\
Options: [CANDIDATE OPTIONS]\\
Answer: [The option index of the answer such as (B)]\\
Explanation: [A concise explanation for the answer]\\
\\
Question: [QUESTION]\\
Options: [CANDIDATE OPTIONS]\\
Answer: [The option index of the answer such as (B)]\\
Explanation: [A concise explanation for the answer]\\
\\
······
\end{tcolorbox}

\begin{tcolorbox}[title = {LaMer: GSM8K}, left = 1mm, right = 1mm, top = 1mm, bottom = 1mm, boxsep = 0mm]
You are an expert in creating math questions. Your goal is to generate a set of new math questions based on the reference fact and examples.
\\\\
Here are the requirements:\\
1. The example should contain a question, a solution to derive the answer to the question, several options (>=3), and an answer.\\
2. Don't make any commonsense mistakes and ensure that your solutions are accurate.\\
3. You SHOULD propose totally new reasoning questions. \\
4. NOTE that the fact should be an implicit guidance for obtaining the true answer, which means the fact SHOULD NOT appear explicitly in the questions, solutions, or options.\\
5. Only one option is the correct answer.\\
6. There are no solutions in the reference examples, you SHOULD generate solutions first and then give the answer for your generated new questions.\\
7. The question could be in any form. You can also add a premise to form a question.\\
\\
Reference:\\
Knowledge: \{\quad\quad\}\\
Examples: \{\quad\quad\}
\\ \\
You MUST generate \{\quad\quad\} new examples. The examples MUST be totally different from each other and the reference examples. Please return your responses in the form:\\
Question: [QUESTION]\\
Solution: [A CONCISE step-by-step SOLUTION to DERIVE the ANSWER to the Question]\\
Options: [CANDIDATE OPTIONS containing the answer]\\
Answer: [The option of the answer such as (B) \$15]
\\\\
Question: [QUESTION] \\
Solution: [A CONCISE step-by-step SOLUTION to DERIVE the ANSWER to the Question]\\
Options: [CANDIDATE OPTIONS containing the answer]\\
Answer: [The option of the answer such as (B) \$15]
\\\\
······
\end{tcolorbox}

\begin{table*}[t]
\centering
\caption{Overall performance of LaMer, LLM2LLM, and baselines. \textbf{Bold} numbers denote the best performance among all methods. \textbf{\textit{Average}} denotes the average performance across all benchmarks.}
\begin{tabular}{lllccccccccc}
\toprule
\textbf{LLMs} & \textbf{Size} & \textbf{Methods}  & \textbf{Label-free} & \textbf{Comm.} & \textbf{AGIEval} & \textbf{ARC} & \textbf{MMLU} & \textbf{BBH} & \textbf{CRASS} & \textbf{GSM-Plus} & \textit{\textbf{Average}} \\ \midrule
 & & Base & - & 67.56 & 32.82 & 74.15 & 49.75 & 28.47 & 71.67 & 29.91 & 50.62 \\
 & & AugGPT & \cellcolor[HTML]{9AFF99}Yes  & 74.53 & 33.03 & 76.93 & 52.56 & 33.82 & 81.67 & 13.78 & 52.33 \\
 & & Naive & \cellcolor[HTML]{9AFF99}Yes  & 67.42 & 33.60 & 72.66 & 50.58 & 32.22 & 80.00 & 31.19 & 52.52 \\
 & & Single & \cellcolor[HTML]{9AFF99}Yes  & 69.47 & 34.87 & 74.85 & 51.44 & 29.53 & 80.00 & 29.03 & 52.74 \\
 & & LLM2LLM & \cellcolor[HTML]{FFCCC9}No & 74.59 & 34.44 & 78.66 & 55.03 & 34.58 & 83.33 & 16.33 & 54.00 \\
\multirow{-6}{*}{\textbf{Mistral}} & \multirow{-5}{*}{7B} & LaMer (Ours) & \cellcolor[HTML]{9AFF99}Yes & 75.10 & 34.50 & 77.22 & 54.52 & 33.72 & 88.33 & 29.23 & \textbf{56.09} \\ \midrule
 & & Base & - & 74.84 & 39.45 & 86.29 & 59.72 & 39.85 & 76.67 & 61.93 & 62.82 \\
 & & AugGPT & \cellcolor[HTML]{9AFF99}Yes  & 76.75 & 38.13 & 86.04 & 60.29 & 36.60 & 76.67 & 14.92 & 55.63 \\
 & & Naive & \cellcolor[HTML]{9AFF99}Yes  & 71.28 & 33.86 & 84.50 & 57.86 & 40.93 & 83.33 & 61.56 & 61.90 \\
 & & Single & \cellcolor[HTML]{9AFF99}Yes & 76.36 & 39.72 & 84.72 & 58.97 & 36.53 & 76.67 & 55.85 & 61.26 \\
 & & LLM2LLM & \cellcolor[HTML]{FFCCC9}No & 75.86 & 36.92 & 87.36 & 60.24 & 39.95 & 88.33 & 56.58 & 63.61 \\
\multirow{-6}{*}{\textbf{LLaMA-3}} & \multirow{-5}{*}{8B} & LaMer (Ours) & \cellcolor[HTML]{9AFF99}Yes  & 78.15 & 40.85 & 86.41 & 60.24 & 39.69 & 88.33 & 59.22 & \textbf{64.70} \\ \midrule
 & & Base & - & 71.88 & 42.10 & 85.48 & 59.53 & 37.58 & 85.00 & 61.79 & 63.34 \\
 & & AugGPT & \cellcolor[HTML]{9AFF99}Yes & 79.41 & 42.71 & 89.78 & 63.66 & 40.00 & 90.00 & 14.75 & 60.04 \\
 & & Naive & \cellcolor[HTML]{9AFF99}Yes & 75.62 & 42.80 & 88.40 & 62.50 & 39.34 & 86.67 & 61.82 & 65.31 \\
 & & Single & \cellcolor[HTML]{9AFF99}Yes & 76.41 & 44.30 & 87.64 & 61.98 & 39.69 & 80.00 & 56.62 & 63.81 \\
 & & LLM2LLM & \cellcolor[HTML]{FFCCC9}No & 80.29 & 43.74 & 89.72 & 63.21 & 40.88 & 80.00 & 60.09 & 65.42 \\
\multirow{-6}{*}{\textbf{Qwen2}} & \multirow{-5}{*}{7B} & LaMer (Ours) & \cellcolor[HTML]{9AFF99}Yes & 78.13 & 45.02 & 88.62 & 62.48 & 40.56 & 86.67 & 61.44 & \textbf{66.13} \\ \midrule
 & & Base & - & 53.26 & 25.80 & 49.97 & 33.73 & 23.29 & 38.33 & 7.08 & 33.07 \\
 & & AugGPT & \cellcolor[HTML]{9AFF99}Yes & 55.42 & 25.76 & 47.73 & 33.69 & 24.31 & 36.67 & 3.95 & 32.50 \\
 & & Naive & \cellcolor[HTML]{9AFF99}Yes & 55.38 & 25.78 & 47.79 & 32.23 & 24.24 & 36.67 & 5.99 & 32.58 \\
 & & Single & \cellcolor[HTML]{9AFF99}Yes & 52.85 & 24.51 & 46.67 & 34.00 & 25.03 & 31.67 & 5.86 & 31.51 \\
 & & LLM2LLM & \cellcolor[HTML]{FFCCC9}No & 54.03 & 25.56 & 49.50 & 35.77 & 25.60 & 40.00 & 7.81 & 34.04 \\
\multirow{-6}{*}{\textbf{Gemma-1.1}} & \multirow{-5}{*}{2B} & LaMer (Ours) & \cellcolor[HTML]{9AFF99}Yes & 55.81 & 25.81 & 48.91 & 34.40 & 25.26 & 41.67 & 6.69 & \textbf{34.08} \\ \bottomrule
\end{tabular}
\label{tab:app_results}
\end{table*}

\section{Prompt for Evaluation}
\label{app:prompts_eval}

\begin{tcolorbox}[title = {Mistral}, left = 1mm, right = 1mm, top = 1mm, bottom = 1mm, boxsep = 0mm]
[INST]Question: \{\quad\quad\} \\
Options: \{\quad\quad\}[/INST]\\
······
\end{tcolorbox}

\begin{tcolorbox}[title = {LLaMA-3 and Qwen2}, left = 1mm, right = 1mm, top = 1mm, bottom = 1mm, boxsep = 0mm]
user\\
Question: \{\quad\quad\}\\
Options: \{\quad\quad\} \\\\
assistant \\
······
\end{tcolorbox}

\begin{tcolorbox}[title = {Gemma-1.1}, left = 1mm, right = 1mm, top = 1mm, bottom = 1mm, boxsep = 0mm]
user\\
Question: \{\quad\quad\}\\
Options: \{\quad\quad\} \\
model \\
······
\end{tcolorbox}

\section{Size of Setected Knowledge Deficiencies}
The size of selected knowledge deficiencies of each LLM in each group can refer to Table~\ref{tab:difi}
\label{app:difi}
\begin{table}[t]
\caption{The size of selected knowledge deficiencies.}
\begin{tabular}{lcccc}
\toprule
LLMs & Easy & Normal & Hard & Unfair \\ \midrule
Mistral & 375 & 375 & 375 & 375 \\
LLaMA-3 & 150 & 150 & 150 & 150 \\
Qwen2 & 375 & 375 & 375 & 375 \\
Gemma-1.1 & 375 & 375 & 375 & 375 \\ \bottomrule
\end{tabular}

\label{tab:difi}
\end{table}

\section{Further Analysis}
\begin{table}
\caption{Statistics of e-CARE}
\centering
\begin{tabular}{lcccc}
\toprule
\textbf{Dataset} & \textbf{Train} & \textbf{Dev} & \textbf{Test} & \textbf{Total} \\ \midrule
e-CARE           & 14,928         & 2,132        & 4,264         & 21,324  \\ \bottomrule  
\end{tabular}

\label{tab:app_preliminary}
\end{table}

\subsection{Deficiency Diagnosis}
\label{app:preliminary}
We formally describe the baselines as follows:
\begin{itemize}
    \item \textbf{Golden Label} adopt the labels to judge the response of a specific LLM, this LLM would answer each question in a chain-of-thought~\cite{wei2022chain} way. If the LLM gives the wrong answers according to the labels on some examples, then the examples are treated as the deficiencies of this LLM. We treat this method as the golden standard for diagnosing the knowledge deficiencies of LLMs.
    \item \textbf{Perplexity} computes the perplexity of each option based on a specific LLM, the option with the lowest perplexity is treated as the answer of the LLM, and then labels are introduced the judge the correctness of the LLM. Similar to Golden Label, wrongly answered examples are treated as the deficiencies of this LLM.
    \item \textbf{Random} method randomly samples the examples from a dataset.
    \item \textbf{Relative Entropy} is the method proposed in this paper.
\end{itemize}

We choose e-CARE~\cite{du2022care} as the dataset for experiments, which is an explainable causal reasoning dataset with two options in each example. The whole set of e-CARE~(train, dev, and test) is adopted for experiments. Statistics of e-CARE can refer to Table~\ref{tab:app_preliminary}. The LLM we used is Mistral-7B-Instruct-v0.2~\cite{jiang2023mistral}.

\subsection{Full results of LLM2LLM}
\label{app:analysis}
We provide the full results of LaMer, all baselines, and LLM2LLM in Table~\ref{tab:app_results}, which demonstrates the advantages of our proposed LaMer.

\end{document}